\documentclass[11pt]{article}

\usepackage[final]{acl}

\usepackage{times}
\usepackage{latexsym}

\usepackage[T1]{fontenc}

\usepackage[utf8]{inputenc}

\usepackage{microtype}

\usepackage{inconsolata}

\usepackage{graphicx}

\usepackage{amsmath} 
\usepackage{amssymb}
\usepackage{algorithm}
\usepackage{algpseudocode}
\usepackage{lipsum}
\usepackage{booktabs}
\usepackage{paralist}
\usepackage{hyperref}
\usepackage{multirow}
\usepackage{makecell}
\usepackage{bm}
\usepackage{tabularx}
\newcolumntype{C}{>{\centering\arraybackslash}X}

\usepackage{pifont} 
\usepackage{xspace}
\usepackage[most]{tcolorbox}
\tcbuselibrary{breakable}
\DeclareTCBListing{promptbox}{ m }{
  colback=gray!5!white,
  colframe=gray!50!black,
  fonttitle=\bfseries\small,
  title={#1},
  breakable,
  left=4pt, right=4pt, top=4pt, bottom=4pt,
  listing only, 
  listing options={
    basicstyle=\ttfamily\footnotesize,
    breaklines=true, 
    columns=fullflexible, 
    showstringspaces=false
  }
}

\newcommand{\name}{\textsc{Mawile}\xspace}

\title{MAWILE: Multi-Axis Workbench for Inspecting LLM Evaluators}

\author{
 \textbf{Jackson Hassell}, 
 \textbf{Farima Fatahi Bayat}, 
 \textbf{Pouya Pezeshkpour}, 
 \textbf{Estevam Hruschka}
\\
 Megagon Labs
\\
    \{jackson, farima, pouya, estevam\}@megagon.ai
\\
}

\begin{document}
\maketitle
\begin{abstract}
Large language model (LLM) judges provide a flexible and scalable method for evaluating model and agent outputs, but their verdicts can be sensitive to incidental changes in the evaluated response, judge instructions, and scoring rubric. Existing systems examine important subsets of these failure modes, but auditing a configured judge requires testing both the judge instrument and the items it evaluates. We introduce \name, a developer-facing workbench for auditing judge sensitivity across four surfaces: the judge prompt, judge rubric, target-system input, and target-system output. Given a user-supplied judge and representative evaluation items, \name constructs and validates controlled perturbations, re-executes the judge, and localizes the resulting sensitivity. Each perturbation declares whether the verdict should remain invariant or change in a specified direction, allowing the same system to measure both robustness to irrelevant variations and sensitivity to meaningful changes. \name audits binary, ordinal, and pairwise judges without requiring gold labels. The code for this tool is available at: \url{github.com/megagonlabs/mawile-judge}.

\end{abstract}
\section{Introduction}
\label{sec:intro}

\begin{figure*}[h]
    \centering
    \includegraphics[width=0.99\textwidth, page=1]{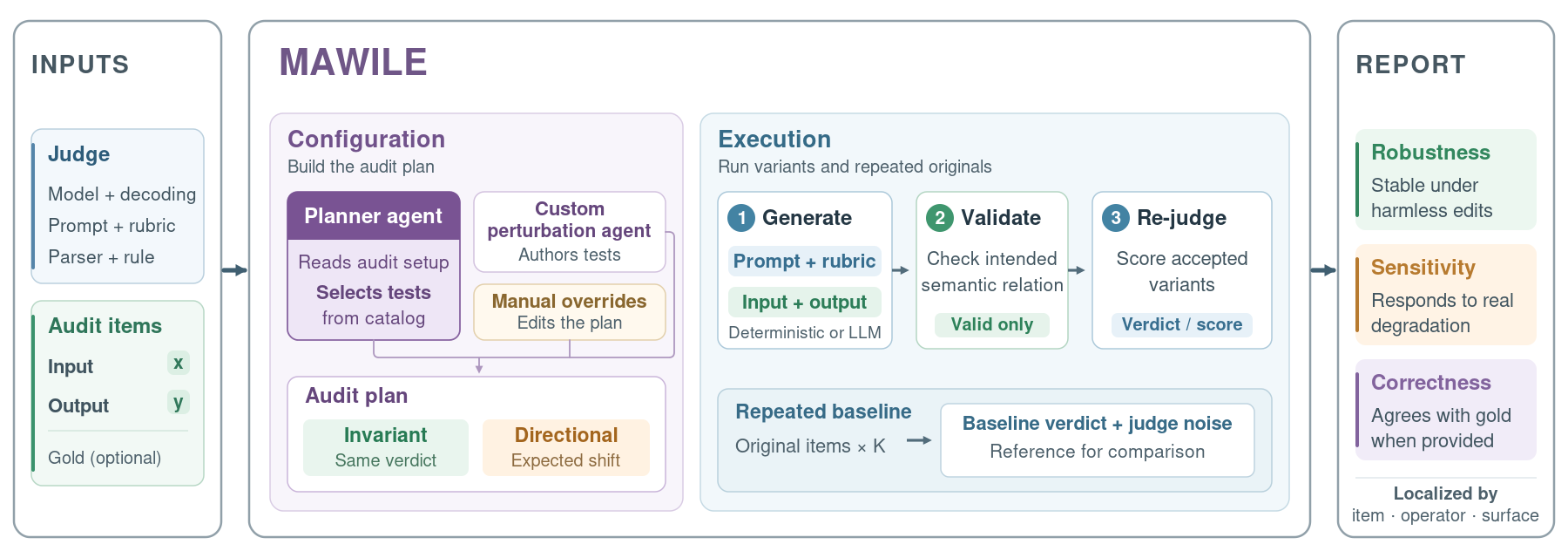}
    \caption{The \name end-to-end judge-auditing workflow. Given a configured judge and representative audit items, the planner agent produces an editable suite of invariant and directional probes. \name then generates and validates these variants, re-executes the judge alongside repeated unmodified baselines, and reports robustness, degradation sensitivity, and---when gold labels are available---correctness.}
    \label{fig:main-diagram}
\end{figure*}

Large language models (LLMs) are increasingly used as flexible, scalable evaluators of model and agent outputs \citep{3666122.3668142, 3692070.3692324, fu-etal-2024-gptscore, pmlr-v267-zhuge25a}. However, their judgments can be sensitive to incidental choices such as candidate order, prompt wording, and rubric presentation \citep{3666122.3668142, bellibatlu2026judgesensebenchmarkpromptsensitivity, 10.1007/978-981-92-0372-7_2}. 
This sensitivity can undermine the reliability of the resulting judgments. While a judge may agree with human labels on a static test set, it may change its decision under semantically irrelevant edits, or fail to adjust when an evaluated output is meaningfully degraded.

Controlled perturbations expose these failures by testing whether a judge remains stable under irrelevant changes and responds appropriately to meaningful ones \citep{ribeiro-etal-2020-beyond, 10.1007/978-981-92-0372-7_2}. We distinguish four perturbation surfaces: the judge rubric, which defines the evaluation criteria and scoring semantics; the judge prompt, which specifies how to perform and return the judgment; the evaluated system's input, such as a user query; and one or more candidate outputs, such as candidate responses to that query. Existing approaches examine complementary subsets of these surfaces, motivating an integrated audit of both the judge instrument and the items it evaluates.

We introduce \name (see Figure \ref{fig:main-diagram}), a developer-facing workbench that integrates these tests into an interactive audit workflow. Given a configured judge and representative evaluation items, a planning agent proposes applicable perturbation operators, which users can edit or extend with custom operators. \name then constructs variants of the selected judge or item fields. Each operator specifies whether the semantic verdict should remain unchanged (invariant) or move in a prescribed direction (directional). A validation stage
checks whether variants satisfy the intended constraints, helping prevent unintended changes from being misinterpreted as judge failures. Finally, \name re-executes the judge and reports sensitivity by evaluation surface, operator, and item, with the underlying evidence available for inspection.

We make the following three contributions. First, we combine item-side and judge-instrument reliability tests within a single end-to-end workflow. Second, we introduce a typed probe interface with explicit applicability, expected-relation, and validation semantics (e.g., context-aware modifications of input-output pairs). Third, we provide an interactive audit process that localizes failures and preserves the evidence needed to inspect and compare judge configurations. \name supports more reliable development and deployment of LLM-based evaluators by making their failure modes easier to identify, diagnose, and compare.

\section{Related Work}
\label{sec:related}

\subsection{Perturbation-Based Evaluation}
\name builds on existing work that associates controlled transformations with an expected relation between the original and transformed predictions \citep{ribeiro-etal-2020-beyond}. 
This principle has been applied to several components of LLM evaluation.
FBI \citep{doddapaneni-etal-2024-finding} injects known quality degradations into evaluated responses to test whether judges detect them, while CALM \citep{ye2025justice} perturbs judge instructions or evaluated responses to quantify a catalog of biases. JudgeSense \citep{bellibatlu2026judgesensebenchmarkpromptsensitivity} 
focuses on decision instability under semantically equivalent paraphrases of the judge prompt.
reWordBench \citep{wu-etal-2025-rewordbench} studies robustness under transformations of reward-model prompts and responses.
Complementary rubric-focused work studies score-rubric ordering, preference drift under natural-language rubric edits, and rubric auditing and repair \citep{10.1007/978-981-92-0372-7_2, ding2026rubrics}.
\name 
unifies these perturbation principles across judge and item fields, with each operator declaring its applicability, expected verdict relation, and validation requirements.

\subsection{Judge Audit Systems}
The Judge Reliability Harness (JRH) \citep{dev2026judgereliabilityharnessstress} generates and validates reliability tests for binary and ordinal judges, aggregates metrics, and supports human review. Its interventions primarily target evaluated responses or agent messages. EvalSense \citep{dejl-pearson-2026-evalsense} provides an interactive framework for configuring and comparing evaluation methods through perturbation-based meta-evaluation. RobustJudge \citep{li2026llmsreliablyjudgeyet} evaluates response attacks and defenses while examining judge models and prompt configurations.

\name audits a configured judge across its prompt, rubric, target-system input, and candidate outputs within one editable workflow. It supports context-aware transformations of individual fields and coupled input--output pairs, and reports failures with their supporting evidence. Table~\ref{tab:related-work} compares the intervention surfaces covered by these systems and related perturbation frameworks.

\begin{table*}[t]
\centering
\small
\begin{tabular}{lcccccc}
\toprule
& \multicolumn{2}{c}{Judge Instrument}
& \multicolumn{3}{c}{Evaluated Instance}
& \multirow{2}{*}{\shortstack{Developer-Facing\\Audit System}} \\
\cmidrule(lr){2-3}
\cmidrule(lr){4-6}
Method
& Rubric $R$
& Prompt $P$
& Input $x$
& Output $y$
& Coupled $x,y$
& \\
\midrule
FBI \citep{doddapaneni-etal-2024-finding}
    & -- & --         & --         & \checkmark & --         & -- \\
reWordBench \citep{wu-etal-2025-rewordbench}
    & -- & --         & \checkmark & \checkmark & \checkmark & -- \\
CALM \citep{ye2025justice}
    & -- & \checkmark & --         & \checkmark & --         & -- \\
JudgeSense \citep{bellibatlu2026judgesensebenchmarkpromptsensitivity}
    & -- & \checkmark         & --         & --         & --         & -- \\
JRH \citep{dev2026judgereliabilityharnessstress}
    & -- & --         & --         & \checkmark & --         & \checkmark \\
EvalSense \citep{dejl-pearson-2026-evalsense}
    & -- & --         & --         & \checkmark & --         & \checkmark \\
RobustJudge \citep{li2026llmsreliablyjudgeyet}
    & \checkmark & \checkmark & -- & \checkmark & --         & \checkmark \\
\midrule
MAWILE
    & \checkmark & \checkmark & \checkmark & \checkmark & \checkmark & \checkmark \\
\bottomrule
\end{tabular}
\caption{
Comparison with prior judge-auditing systems and perturbation frameworks. A checkmark indicates that the method automatically constructs or applies controlled variants of the corresponding surface as an intervention target, not that a given surface is merely configurable.}
\label{tab:related-work}
\end{table*}
\section{Methodology}
\label{sec:method}

\name audits a user-supplied LLM judge by applying controlled changes to the judge instrument and the items it evaluates. As shown in Figure~\ref{fig:main-diagram}, an audit begins with a configured judge and a representative collection of evaluation items. 
\name first establishes the judge's baseline behavior and 
estimates judge consistency
by repeatedly evaluating the unmodified items. It then 
selects applicable perturbations, constructs and validates the corresponding variants,
re-executes the judge, and checks whether verdicts remain stable under invariant perturbations or move in the expected direction under directional perturbations.
The resulting report localizes sensitivity by evaluation surface, perturbation type, and item. 

\subsection{Audit Setup}
\label{sec:audit-setup}

We represent a configured judge as
\(\mathcal{J}=(M,P,R,\kappa,\mathcal{Z})\), where \(M\) specifies the judge model and decoding configuration, \(P\) is the evaluation prompt, \(R\) is the rubric, the judge harness \(\kappa\) processes the raw judge response, and \(\mathcal{Z}\) 
is the verdict space.
An evaluation item contains a target-system input \(x_i\) and one or more target-system outputs \(y_i\). A judge call produces
\begin{equation}
z_i = \kappa\bigl(M(P,R,x_i,y_i)\bigr), \qquad z_i \in \mathcal{Z}.
\label{eq:judge}
\end{equation}
This decomposition exposes four perturbation surfaces: the judge prompt \(P\), judge rubric \(R\), target-system input \(x_i\), and target-system output(s) \(y_i\). 
These four surfaces are varied within an audit, while \(M\), \(\kappa\), and \(\mathcal{Z}\) are held fixed.

The verdict
space \(\mathcal{Z}\) depends on the type of judge. Binary judges specify pass or fail, ordinal judges rank an output along a specified score scale, and pairwise judges choose which of two options is the better response to the input.

Before applying perturbations, \name repeatedly evaluates each original item. These runs establish a baseline verdict \(b_i\), defined as the modal class for categorical judges and the mean score for numeric judges. \name also measures disagreement across repeated calls to quantify the judge's intrinsic stochasticity. Together, these statistics help distinguish perturbation-induced sensitivity from ordinary run-to-run variation.

\subsection{Typed Perturbation Operators}
\label{sec:typed-probes}

A perturbation operator perturbs one or more of the four audit surfaces.
Each operator declares its target surface, applicability conditions, generation procedure, validation procedure, and the expected relation between the original and perturbed verdicts as shown in Figure \ref{fig:main-diagram}. \name supports two relation types. \textit{Invariant} operators introduce perturbations that should not affect the semantic verdict, such as swapping the order of two candidates in pairwise evaluation.
\textit{Directional} operators, conversely, introduce perturbations with a known expected effect on the verdict. The operator specifies both the intended direction and the orientation of the judge's output scale. For example, adding a new requirement to \(x_i\) while leaving the corresponding output unchanged should lower the evaluation score.

Directional operators are restricted to binary and ordinal judges. 
In pairwise evaluation, degrading one candidate does not necessarily imply a preference reversal; establishing such an expectation would require verifying its quality relative to the unchanged candidate.
For pointwise judges, directional perturbations are only applied to positive items (for binary judges) or items not already at the bottom of the scale (for ordinal judges).

Together, these two relation types test two complementary properties. Invariant probes measure robustness to changes that should preserve the underlying decision, while directional probes measure whether the judge responds to changes that should alter it. 

\subsection{Audit Planning and Perturbation Construction}
\label{sec:audit-planning}

\name provides a catalog of perturbation operators spanning the four evaluation surfaces. The catalog includes deterministic transformations, such as candidate-order swaps and structure-preserving formatting changes, as well as LLM-generated transformations, such as semantic paraphrases, controlled degradations, and task-dependent modifications (see Tables \ref{tab:perturbation-catalog-invariant} and \ref{tab:perturbation-catalog-directional} for the full included perturbation catalog).
Users may select operators directly, add custom operators, or ask a planner agent to propose an audit based on the judge configuration and representative dataset items. The proposed plan remains editable before execution, keeping task-specific requirements under user control.

LLM-generated perturbations receive the context needed to preserve a coherent evaluation instance. When modifying a target-system output $y_i$, the generator also observes its corresponding input $x_i$, and vice versa. 
We use this context-aware construction to avoid attributing judge failures to perturbations that instead create incoherent or implausible input--output pairs.
Operators may also modify both $x_i$ and $y_i$ jointly, enabling transformations such as consistent entity renaming to probe potential social biases in the judge. 

Validation checks whether each variant satisfies its operator-specific requirements. Some operators, such as candidate-position swaps, satisfy their constraints by construction, as they preserve both candidate content and identity. Variants requiring semantic assessment are instead checked by a separate validation agent. Invariant perturbations must preserve the properties relevant to the judgment, while directional perturbations must introduce the specified degradation. Rejected variants are excluded from judge execution. The generator, validator, and judge are independently configurable roles.

\subsection{Report Generation}
\label{sec:report}

The final report aggregates results by evaluation surface, operator, and item. Users can inspect the original and perturbed fields, canonical judge outputs, validation outcomes, 
and violations of each operator's expected behavior, such as a preference change after a position swap or failure to lower a score after a validated degradation.
This supports analysis of both global sensitivity trends and local inspection of the items driving them. If gold labels are available, the report also includes judge accuracy.
In addition to automatically computed metrics, \name provides an LLM-generated summary of key findings and recommended follow-up actions to improve judge reliability and robustness (Figure~\ref{fig:walkthrough_report_overview}).

\begin{figure*}[h]
    \centering
    \includegraphics[width=0.99\textwidth, page=1]{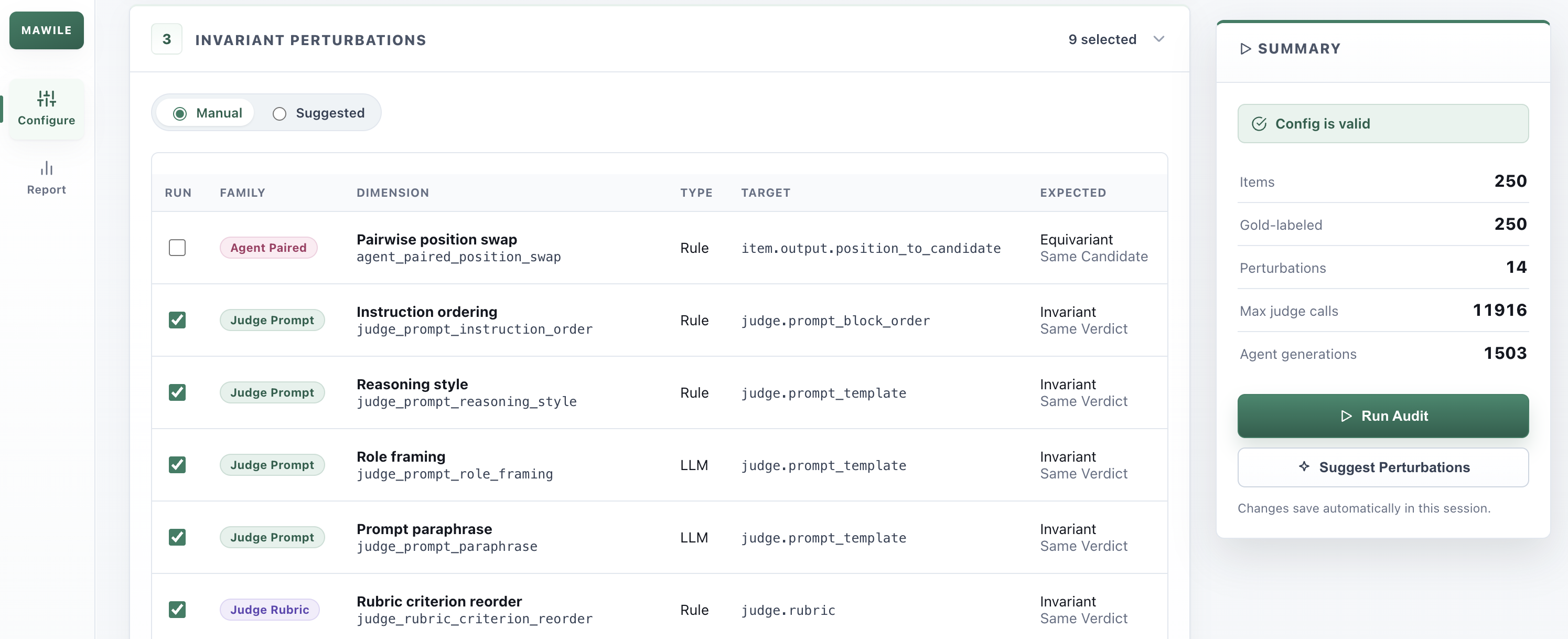}
    \caption{Screenshot of the \name Web UI's editable perturbation catalog. During configuration, developers can inspect each operator's family, construction method, target field, and expected verdict relation, then enable or disable it before execution. A planner agent can also suggest perturbations tailored to the dataset.}
    \label{fig:walkthrough_cropped_config}
\end{figure*}

\begin{figure*}[h]
    \centering
    \includegraphics[width=0.99\textwidth, page=1]{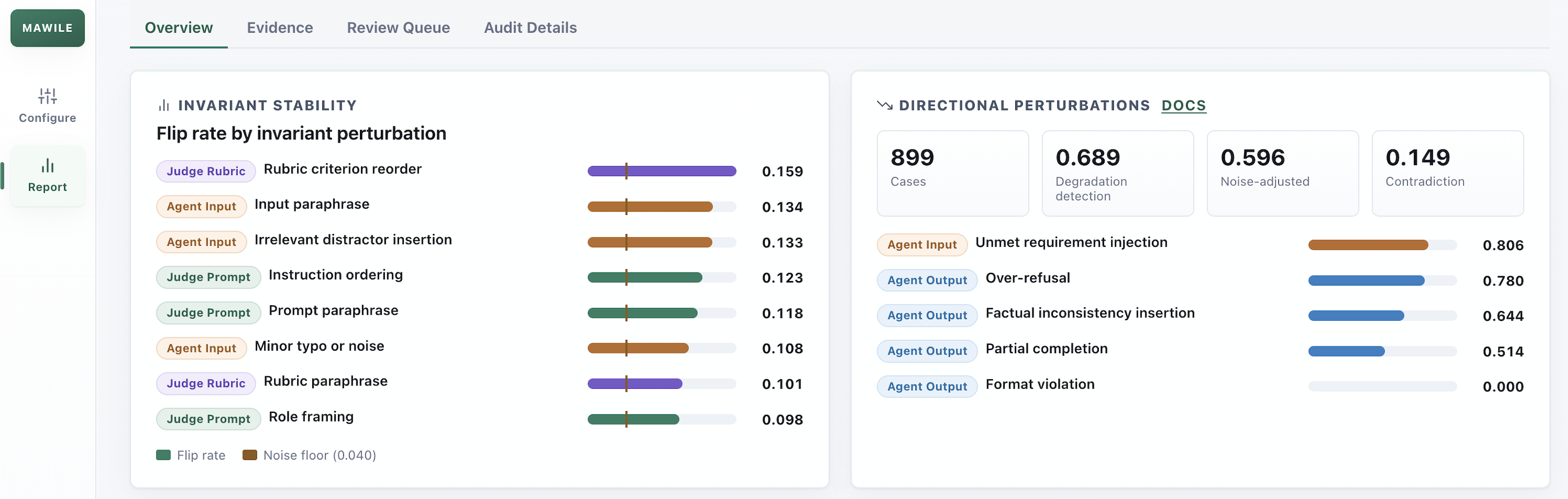}
    \caption{Screenshot of the \name web UI's per-operator results for the MT-Bench walkthrough using DeepSeek-V4-Flash as the judge and Kimi K3 as the perturbation generator.}
    \label{fig:walkthrough_cropped_report}
\end{figure*}

\section{User Interface Walkthrough}

We illustrate the \name system through an audit of DeepSeek-V4-Flash on 250 MT-Bench items (the same items used in Table \ref{tab:main-results}), using Kimi K3 to generate perturbations. 

\paragraph{Configuration.}
During the configuration step, the developer 
can set or override audit hyperparameters. In this experiment, the developer defined a rubric that grades each item on a 1--10 scale 
, allows a one-point change to count as invariant, and treats any score decrease as a valid directional degradation.
The planning agent proposed nine invariant and five directional operators, along with rationales for each choice. For example, the planner excluded verbosity shifts from the invariant suite because 
response depth is explicitly included in the rubric and may legitimately affect the score.
It includes rubric reordering, which changes the presentation of the scoring criteria while preserving their content (see Figure \ref{fig:walkthrough_config_perturbation_catalog} for the full justification). In the editable catalog (Figure~\ref{fig:walkthrough_cropped_config}), the developer can review the selected operators and their expected effects, and enable or disable each operator.

\paragraph{Generated Report.}
\name constructed and validated the perturbations, then evaluated the accepted variants against the repeated baseline. The report combines aggregate metrics, an LLM-generated summary, per-perturbation results, and items flagged for manual review. Figure \ref{fig:walkthrough_cropped_report} (left) shows that rubric reordering produces the highest invariant flip rate, at 15.9\% compared with the 4.0\% baseline flip rate on unmodified items. 
The right panel shows that the judge is completely unable to detect the format-violation errors, but is substantially more sensitive to injected unmet requirements.

\paragraph{Evidence Inspection.}
The developer can also examine the examples underlying these reported findings. In this experiment, the generated summary highlighted an item whose score drops from 9.5 to 2.0 (averaged across multiple calls) after rubric reordering. The developer can review this case in the Review Queue tab (Figure \ref{fig:walkthrough_report_review_queue}) to inspect that item manually.
Under the Evidence tab, they can inspect examples of perturbed questions to verify that the results are coherent in context. 
This supports both rapid interpretation of aggregate findings and deeper verification of the evidence behind them.

\begin{table*}[t]
\centering
\small
\setlength{\tabcolsep}{3pt}
\begin{tabular}{@{}lllcccc@{}}
\toprule
& & & \multicolumn{2}{c}{Baseline}
& \multicolumn{2}{c}{Perturbation Response} \\
\cmidrule(lr){4-5}
\cmidrule(l){6-7}
Dataset
& \makecell{Perturbation\\Generator}
& Judge
& \makecell{Accuracy $\uparrow$}
& \makecell{Repeat\\Flip Rate $\downarrow$}
& \makecell{Invariant\\Flip Rate $\downarrow$}
& \makecell{Degradation\\Detection Rate $\uparrow$} \\
\midrule
\multirow{6}{*}{\makecell[l]{MT-Bench\\{\scriptsize Ordinal}}}
 & \multirow{3}{*}{Qwen 3.8 Max}
 & GPT-5.6 Luna & 56.0 & 2.3 & 6.0 & 62.9 \\
 & & DeepSeek-V4-Flash
 & \textbf{59.1} & 5.3 & 11.7 & \underline{65.4} \\
 & & Gemma 3 4B
 & 51.2 & \underline{0.5} & \underline{2.8} & 47.4 \\
\cmidrule(l){2-7}
 & \multirow{3}{*}{Kimi K3}
 & GPT-5.6 Luna & 53.6 & 3.2 & 5.7 & \underline{65.4} \\
 & & DeepSeek-V4-Flash
 & \underline{59.0} & 4.0 & 12.2 & \textbf{68.9} \\
 & & Gemma 3 4B
 & 49.6 & \textbf{0.1} & \textbf{2.4} & 53.0 \\
\midrule
\multirow{6}{*}{\makecell[l]{Search Arena\\{\scriptsize Pairwise}}}
 & \multirow{3}{*}{Qwen 3.8 Max}
 & GPT-5.6 Luna & 62.0 & 5.3 & 10.5 & -- \\
 & & DeepSeek-V4-Flash
 & \underline{64.1} & 1.6 & \underline{8.6} & -- \\
 & & Gemma 3 4B
 & 53.3 & \textbf{0.3} & 16.0 & -- \\
\cmidrule(l){2-7}
 & \multirow{3}{*}{Kimi K3}
 & GPT-5.6 Luna & 58.0 & 5.2 & 10.5 & -- \\
 & & DeepSeek-V4-Flash
 & \textbf{66.1} & 2.6 & \textbf{8.1} & -- \\
 & & Gemma 3 4B
 & 53.3 & \underline{0.4} & 13.8 & -- \\
\midrule
\multirow{6}{*}{\makecell[l]{GSM8K\\{\scriptsize Binary}}}
 & \multirow{3}{*}{Qwen 3.8 Max}
 & GPT-5.6 Luna & 96.8 & \textbf{0.1} & 0.7 & 41.1 \\
 & & DeepSeek-V4-Flash
 & \textbf{98.4} & \underline{0.3} & \textbf{0.3} & 48.6 \\
 & & Gemma 3 4B
 & 90.4 & 0.7 & 3.9 & 10.0 \\
\cmidrule(l){2-7}
 & \multirow{3}{*}{Kimi K3}
 & GPT-5.6 Luna
 & 96.0 & \textbf{0.1} & 0.7 & \underline{54.9} \\
 & & DeepSeek-V4-Flash
 & \underline{98.0} & \underline{0.3}
 & \underline{0.6} & \textbf{67.6} \\
 & & Gemma 3 4B
 & 90.8 & 0.4 & 3.6 & 9.2 \\
\bottomrule
\end{tabular}
\caption{Aggregate audit results for ordinal, pairwise, and binary judges across two perturbation generators and three judge models. Within each dataset, bold and underlining mark the best and second-best judges in each metric, respectively. All numbers are percentages. For each dataset, a Kimi K3 planner selected one operator suite, which is held fixed across all six generator-judge combinations.}
\label{tab:main-results}
\end{table*}

\section{Experimental Setup}
\subsection{Datasets}

We evaluate \name on three datasets chosen to span common judge output types, evaluation domains, and interaction settings, ranging from scalar scoring to pairwise preference and binary correctness judgments. 
In \textbf{MT-Bench} \cite{3666122.3668142}, the judge grades an agent's multi-turn conversation with a user on a rubric of 1-10.
In \textbf{Search Arena} \cite{miroyan2026search}, given two agentic responses (with traces) to a user query that requires searching the internet, the judge evaluates which of the two is better.
In \textbf{GSM8K} \cite{cobbe2021trainingverifierssolvemath}, given a math problem and an agent's answer and justification, the judge decides if the agent's answer and decision process is correct or not.

\subsection{Model and Perturbation Setup}
We run a full $3\times3\times2$ comparison across three datasets, using GPT-5.6 Luna \citep{openai2026gpt56}, DeepSeek-V4-Flash \citep{deepseek2026v4}, and Gemma 3 4B \citep{gemma2025gemma3} as judges, and Qwen3.8-Max \citep{qwen2026qwen38} and Kimi K3 \citep{kimi2026k3} as perturbation generators.
To isolate the effect of the perturbation generator, Kimi K3 selects one fixed set of perturbation operators for each dataset. The same operator set is then used for every judge--generator combination.

\subsection{Metrics}
We report four aggregate metrics. \emph{Accuracy} measures agreement between the baseline judgment and the provided correct answer. \emph{Repeat flip rate} measures disagreement across repeated calls on the unmodified item, capturing the judge's intrinsic instability. \emph{Invariant flip rate} measures how often an accepted meaning-preserving perturbation changes the judgment beyond the permitted tolerance. Lastly, \emph{degradation detection rate} measures how often a directional perturbation moves the judgment in the expected direction. 

\section{Experimental Results}
\label{sec:exp}

\paragraph{Accuracy does not imply robustness.}
Across all six dataset--generator settings, DeepSeek-V4-Flash achieves the highest accuracy, followed by GPT-5.6 Luna and Gemma 3 4B. The same ordering holds for degradation detection wherever directional probes are applicable, suggesting that more accurate judges are also generally better at recognizing meaningful degradations. 
Invariant robustness, however, is task-dependent: on MT-Bench, Gemma 3 4B has the lowest invariant flip rates (2.4--2.8pp), followed by GPT-5.6 Luna (5.7--6.0pp) and DeepSeek-V4-Flash (11.7--12.2pp), while on Search Arena and GSM8K DeepSeek-V4-Flash is most robust and Gemma 3 4B least robust. Thus, robustness to meaning-preserving changes cannot be inferred from accuracy alone.

\paragraph{Repeatability does not imply perturbation robustness.}
On Search Arena, Gemma 3 4B exhibits only 0.3--0.4\% repeat
flips on unmodified items, but 13.8--16.0\% invariant flips
after meaning-preserving perturbations.
Similar gaps elsewhere show that perturbation sensitivity captures failures beyond ordinary run-to-run stochasticity.

\paragraph{Directional probes are more sensitive to the perturbation generator.}
Switching between Qwen 3.8 Max and Kimi K3 yields mean absolute differences of 0.5 percentage points in invariant flip rate and 7.5 points in degradation detection.
Kimi K3 improves detection for all judges on MT-Bench and for GPT-5.6 Luna and DeepSeek-V4-Flash on GSM8K, suggesting that generator choice matters more for constructing meaningful degradations than invariant variants.

\section{Conclusion}
\label{sec:conclusion}

We introduced \name, a developer-facing system for auditing this distinction through validated perturbations to the judge prompt, rubric, evaluated input, and evaluated output. Across ordinal, pairwise, and binary evaluations, our experiments show that judge sensitivity depends on both the model and task.
Higher accuracy is often associated with better detection of meaningful degradations, but does not necessarily imply greater robustness to irrelevant perturbations.
These findings motivate evaluating judge accuracy alongside repeatability and perturbation sensitivity. By localizing failures to individual operators, items, and evaluation surfaces, \name provides an inspectable basis for diagnosing unreliable judge behavior, including in settings where gold labels are unavailable.

\section*{Limitations}

\name measures judge sensitivity, not necessarily judge correctness. A judge that remains stable under every tested perturbation may still apply an incorrect rubric or reproduce systematic biases. Gold labels can reveal some such failures when available, but label-free audits cannot certify that a stable verdict is valid.

The conclusions of an audit also depend on the coverage and validity of its perturbations. The operator catalog represents only a subset of possible real-world variation (though we allow both manual and LLM-authored custom perturbations to mitigate this), while LLM-generated perturbations and their validators may introduce their own errors or biases. Passing the resulting suite is therefore evidence of reliability under the tested conditions, rather than a general guarantee. 

\bibliography{custom}

\appendix
\section{Appendix}

\subsection{Additional Walkthrough Screenshots}

\begin{figure*}[h]
    \centering
    \includegraphics[width=0.99\textwidth, page=1]{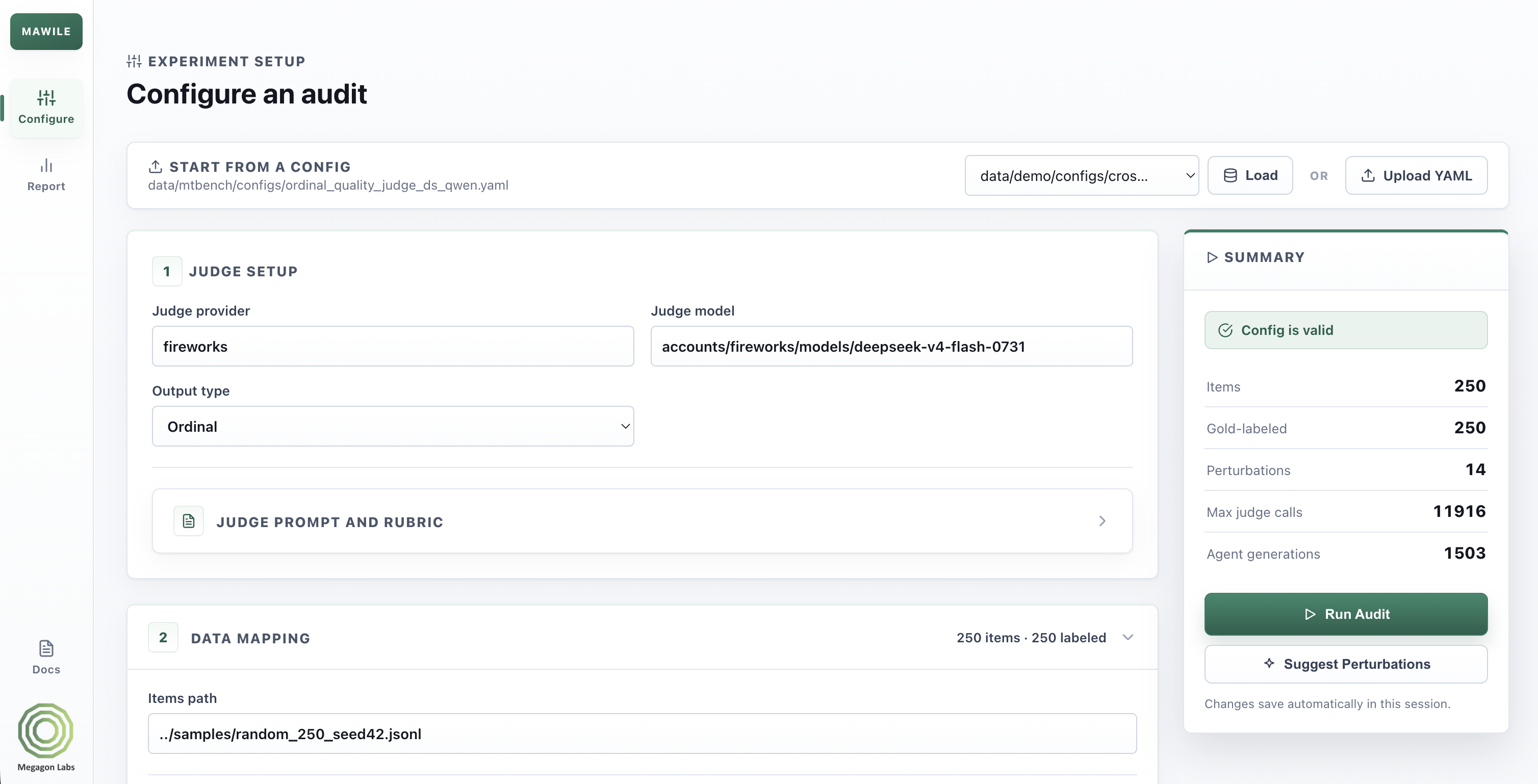}
    \caption{Screenshot of the \name web UI for configuring a judge and audit dataset.}
    \label{fig:walkthrough_config_loading}
\end{figure*}

\begin{figure*}[h]
    \centering
    \includegraphics[width=0.99\textwidth, page=1]{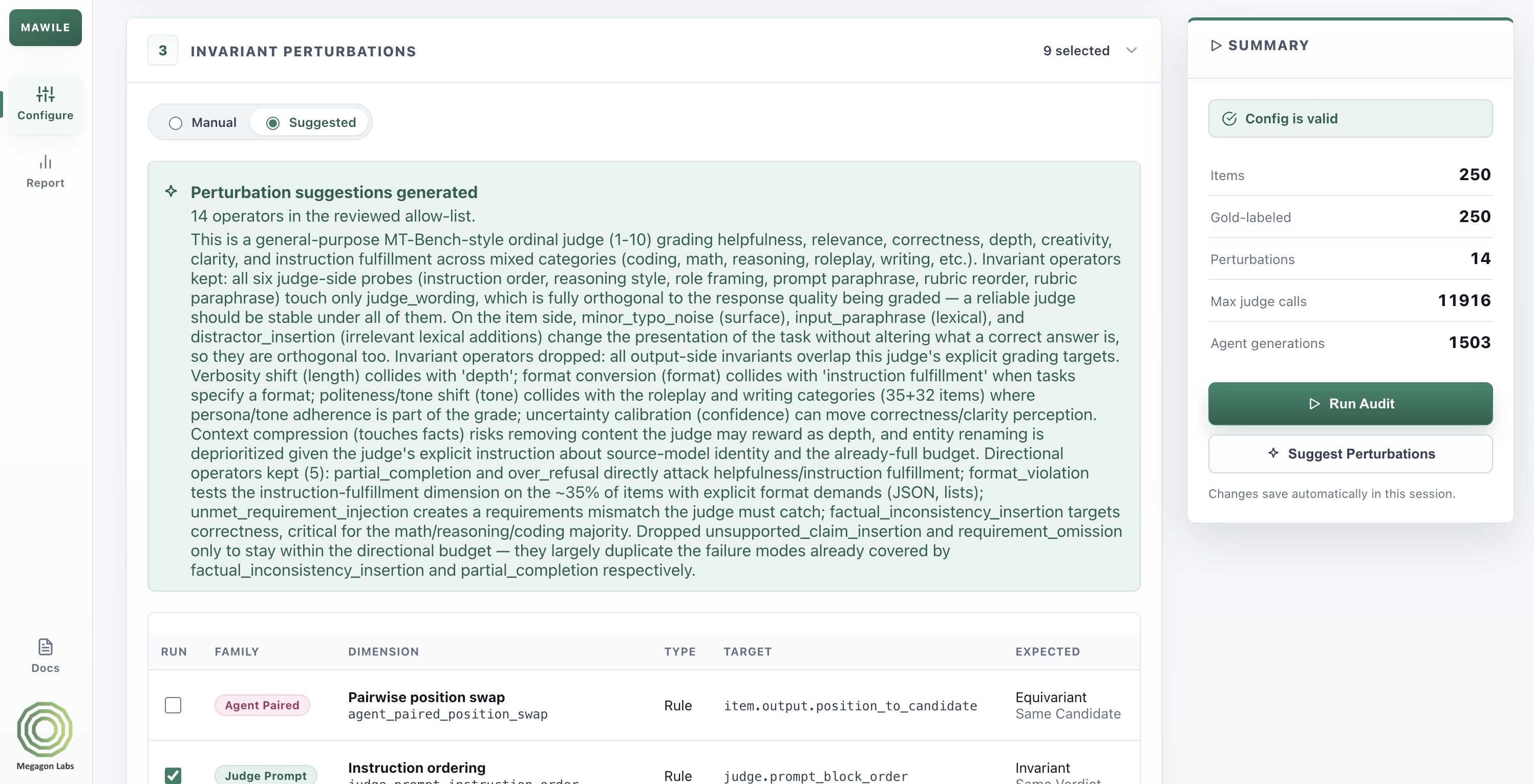}
    \caption{Screenshot of the \name planning agent's proposed perturbation suite for the MT-Bench walkthrough, with justification.}
    \label{fig:walkthrough_config_perturbation_catalog}
\end{figure*}

\begin{figure*}[h]
    \centering
    \includegraphics[width=0.99\textwidth, page=1]{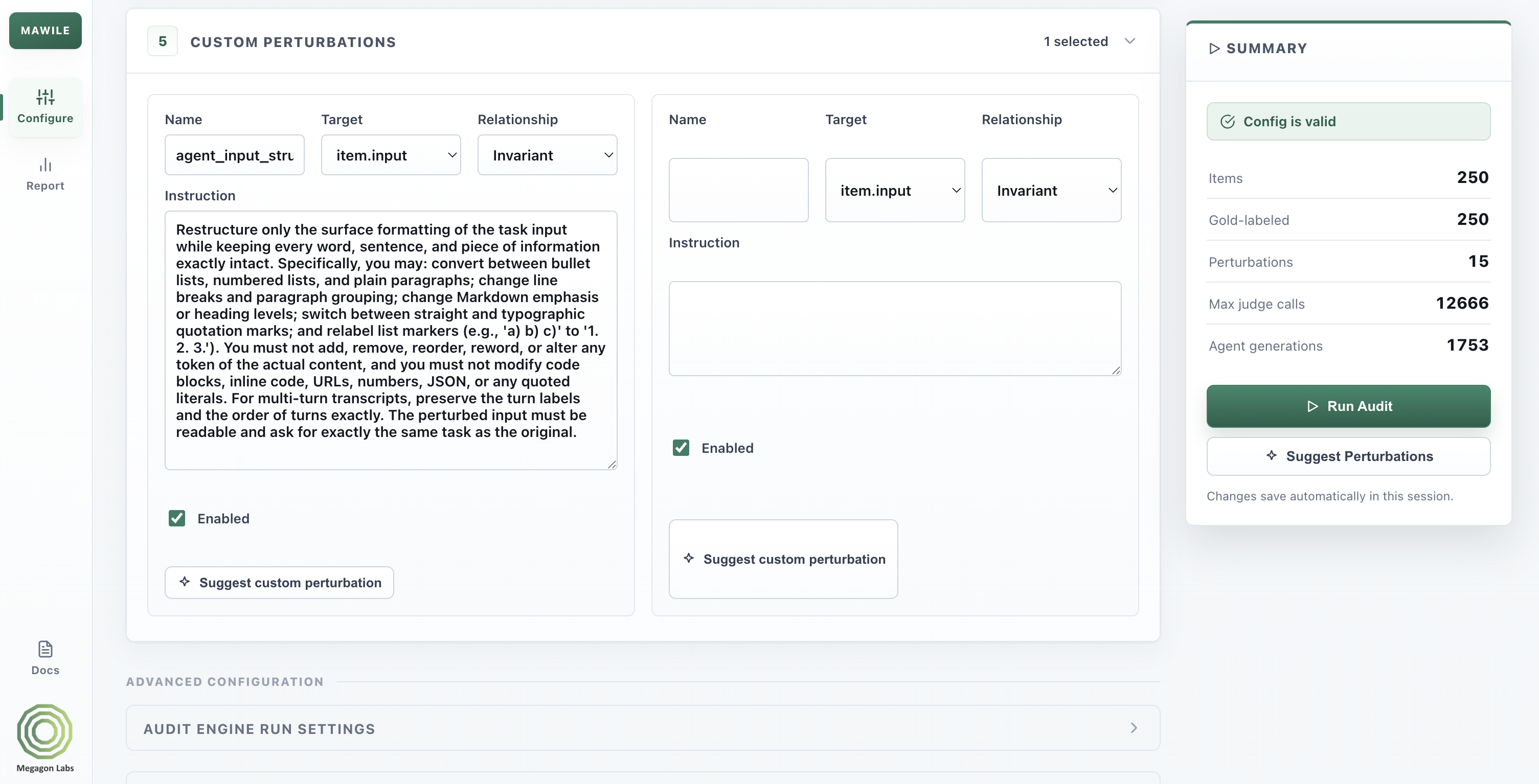}
    \caption{Screenshot of the \name web UI for defining a custom perturbation, with an example of an LLM-suggested custom perturbation.}
    \label{fig:walkthrough_config_custom_perturbations}
\end{figure*}

\begin{figure*}[h]
    \centering
    \includegraphics[width=0.99\textwidth, page=1]{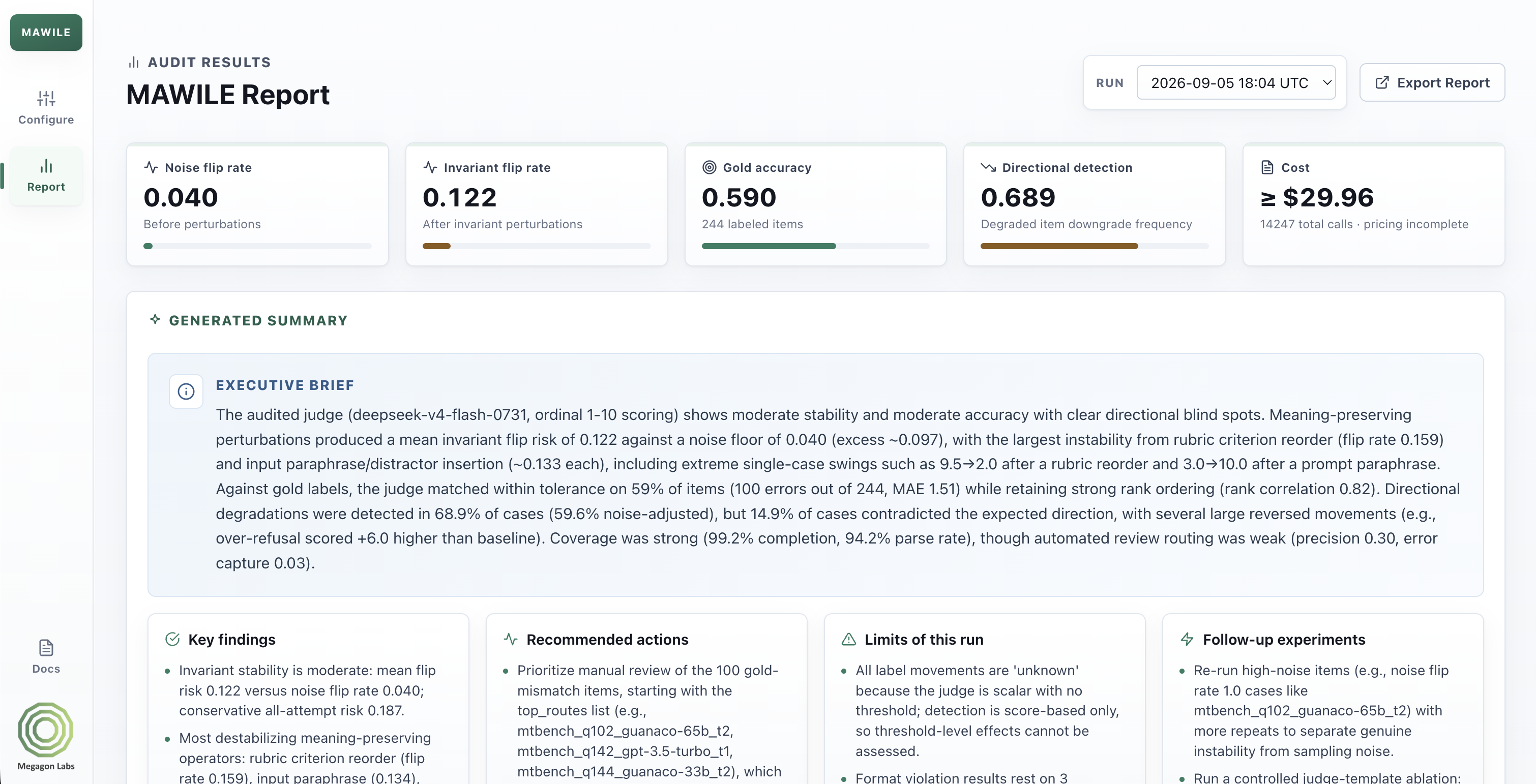}
    \caption{Screenshot of the \name report overview for the MT-Bench walkthrough.}
    \label{fig:walkthrough_report_overview}
\end{figure*}

\begin{figure*}[h]
    \centering
    \includegraphics[width=0.99\textwidth, page=1]{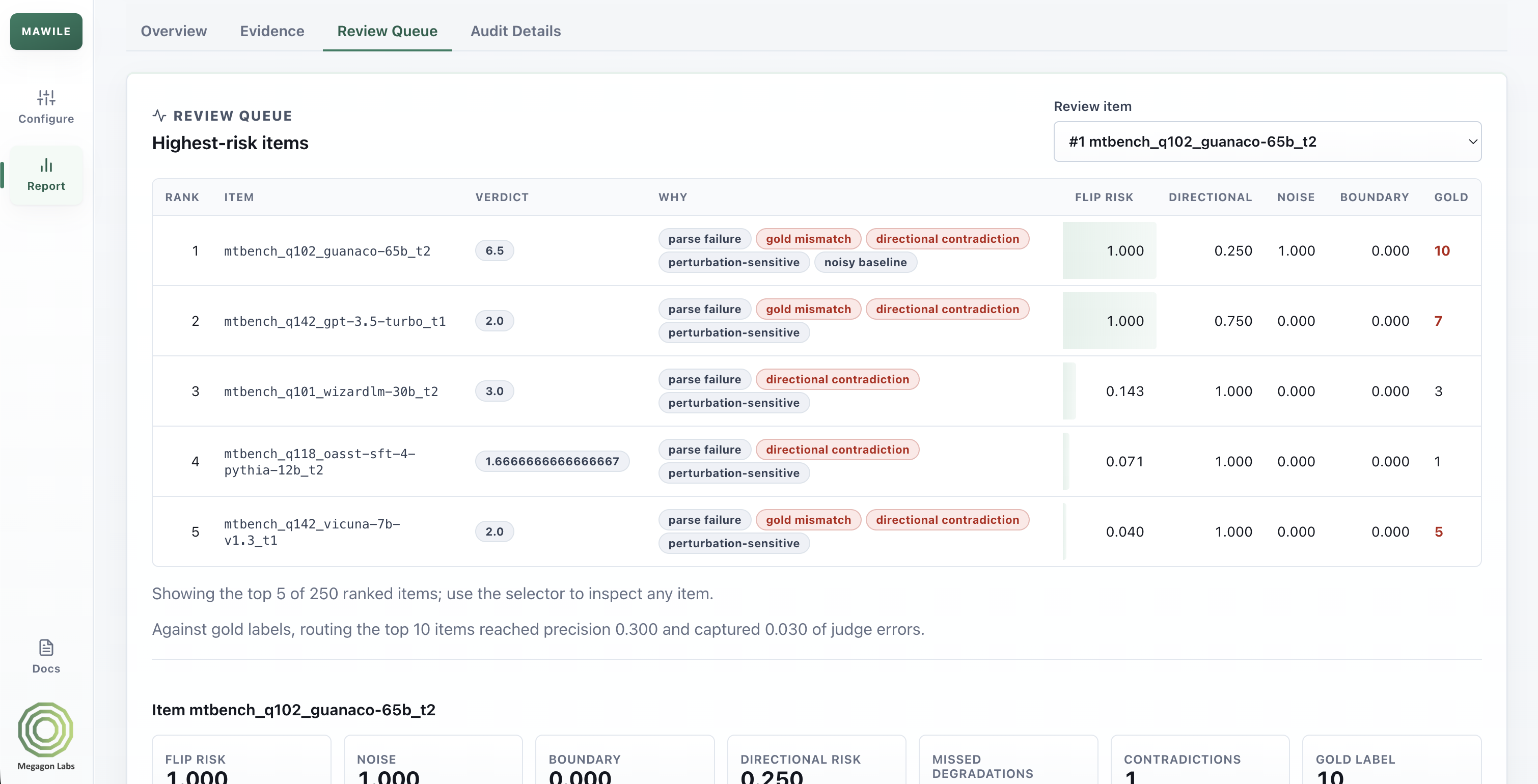}
    \caption{Screenshot of the \name review queue for inspecting flagged items from the MT-Bench walkthrough.}    
    \label{fig:walkthrough_report_review_queue}
\end{figure*}

Figures \ref{fig:walkthrough_config_loading}, \ref{fig:walkthrough_config_perturbation_catalog}, \ref{fig:walkthrough_config_custom_perturbations}, \ref{fig:walkthrough_report_overview}, and \ref{fig:walkthrough_report_review_queue} supplement the MT-Bench walkthrough with additional views of audit configuration and result inspection. They show the judge and dataset settings, the planner's justifications, custom perturbation authoring, and the report overview and review queue. These interfaces allow developers to inspect and revise the proposed perturbation suite before execution, then examine the individual items underlying the reported findings.

\subsection{Decoding Parameters}

All models used reasoning medium effort and default temperature for all calls. 

\subsection{Perturbation Catalog}
\label{app:perturbation-catalog}

The built-in catalog contains 16 invariant operators and 7 directional degradation operators (Tables~\ref{tab:perturbation-catalog-invariant} and~\ref{tab:perturbation-catalog-directional}). Invariant probes should preserve the verdict; directional probes intentionally make an item worse and should therefore elicit a worse verdict. ``Rule'' denotes a deterministic edit and ``LLM'' a perturbation-model rewrite. Semantic validation is independent of the judge under test and applies to both LLM rewrites and rule-based edits when needed.
Judge-level edits apply globally while item-level edits are applied each item independently.

\begin{table*}[t]
\centering
\small
\setlength{\tabcolsep}{4pt}
\renewcommand{\arraystretch}{1.08}
\begin{tabular}{@{}>{\raggedright\arraybackslash}p{\dimexpr0.15\textwidth-1.5\tabcolsep\relax}>{\raggedright\arraybackslash}p{\dimexpr0.29\textwidth-1.5\tabcolsep\relax}>{\centering\arraybackslash}p{\dimexpr0.08\textwidth-1.5\tabcolsep\relax}>{\raggedright\arraybackslash}p{\dimexpr0.48\textwidth-1.5\tabcolsep\relax}@{}}
\toprule
\multicolumn{4}{@{}l}{\textbf{Invariant Perturbation Catalog}} \\
\addlinespace[2pt]
Surface & Perturbation & Gen. & Description \\
\midrule
Judge prompt & Instruction ordering & Rule & Moves output-format instructions before the prompt and rubric. \\
 & Reasoning style & LLM & Requests direct, step-by-step, or criterion-by-criterion reasoning. \\
 & Role framing & LLM & Changes the evaluator persona without changing the task. \\
 & Prompt paraphrase & LLM & Rewords the judge instructions while preserving their meaning. \\
\addlinespace[3pt]
Judge rubric & Rubric criterion reorder & Rule & Reorders independent rubric blocks without changing their content. \\
 & Rubric paraphrase & LLM & Rewords scoring criteria while preserving their meaning and thresholds. \\
\addlinespace[3pt]
Agent input & Minor typo or noise & LLM & Introduces minor prose typos without changing task meaning. \\
 & Input paraphrase & LLM & Rewords the input while preserving the task and requirements. \\
 & Irrelevant distractor insertion & LLM & Adds plausible but irrelevant context to the input. \\
 & Context compression & LLM & Removes redundant input context while retaining needed information. \\
\addlinespace[3pt]
Agent output & Output format conversion & LLM & Changes the response format while preserving the answer. \\
 & Politeness or tone shift & LLM & Changes response tone while preserving its substance. \\
 & Uncertainty calibration & LLM & Changes expressed confidence without changing the answer. \\
 & Verbosity shift & LLM & Shortens or lengthens the response without changing substantive claims. \\
\addlinespace[3pt]
Agent paired & Pairwise position swap & Rule & Swaps candidate display positions while preserving candidate identities. \\
 & Entity renaming & LLM & Renames entities consistently across the input and output. \\
\bottomrule
\end{tabular}
\caption{Built-in invariant perturbations.}
\label{tab:perturbation-catalog-invariant}
\end{table*}

\begin{table*}[t]
\centering
\small
\setlength{\tabcolsep}{4pt}
\renewcommand{\arraystretch}{1.08}
\begin{tabular}{@{}>{\raggedright\arraybackslash}p{\dimexpr0.15\textwidth-1.5\tabcolsep\relax}>{\raggedright\arraybackslash}p{\dimexpr0.29\textwidth-1.5\tabcolsep\relax}>{\centering\arraybackslash}p{\dimexpr0.08\textwidth-1.5\tabcolsep\relax}>{\raggedright\arraybackslash}p{\dimexpr0.48\textwidth-1.5\tabcolsep\relax}@{}}
\toprule
\multicolumn{4}{@{}l}{\textbf{Directional Perturbation Catalog}} \\
\addlinespace[2pt]
Surface & Perturbation & Gen. & Description \\
\midrule
Agent input & Unmet requirement injection & LLM & Adds a task requirement that the existing response does not satisfy. \\
\addlinespace[3pt]
Agent output & Partial completion & Rule & Removes the final separable part of a response. \\
 & Format violation & LLM & Breaks one previously satisfied output-format requirement. \\
 & Unsupported claim insertion & LLM & Adds one unsupported substantive claim to the response. \\
 & Factual inconsistency insertion & LLM & Introduces one factual or reasoning inconsistency. \\
 & Requirement omission & LLM & Removes or weakens content that satisfies a task requirement. \\
 & Over-refusal & LLM & Replaces an answer to an answerable task with an unnecessary refusal. \\
\bottomrule
\end{tabular}
\caption{Built-in directional degradation perturbations. Directional probes are available only for pointwise audits.}
\label{tab:perturbation-catalog-directional}
\end{table*}

\end{document}